\documentclass[10pt,twocolumn,letterpaper]{article}

\usepackage[camera-ready]{cvpr}      
\usepackage{float}
\usepackage{amsmath}
\renewcommand{\and}{\hspace{0.6cm}}

\usepackage[dvipsnames]{xcolor}

\renewcommand{\bold}[1]{\mbox{\boldmath$#1$}}

\renewcommand{\vec}[1]{{\bold #1}} 

\newcommand{\mq}[1]{{\mbox{{\sffamily{#1}}}}}

\newcommand{\RR}{\mathbb{R}}

\newcommand{\norm}[1]{\lVert#1\lVert}

\definecolor{cvprblue}{rgb}{0.21,0.49,0.74}
\usepackage[pagebackref,breaklinks,colorlinks,citecolor=cvprblue]{hyperref}

\def\paperID{12902} 
\def\confName{CVPR}
\def\confYear{2024}

\title{3D LiDAR Mapping in Dynamic Environments\\Using a 4D Implicit Neural Representation}

\author{
Xingguang Zhong$^1$
\and
Yue Pan$^1$
\and
Cyrill Stachniss$^{1,2}$
\and
Jens Behley$^1$\\
{\small $^1$Center for Robotics, University of Bonn, $^2$Lamarr Institute for Machine Learning and Artificial Intelligence}\\
{\tt\small \{zhong, yue.pan, cyrill.stachniss, jens.behley\}@igg.uni-bonn.de}
}\vspace{-0.1cm}

\begin{document}

\maketitle

\begin{abstract}
  Building accurate maps is a key building block to enable reliable localization, planning, and navigation of autonomous vehicles.
  We propose a novel approach for building accurate maps of dynamic environments utilizing a sequence of LiDAR scans.
  To this end, we propose encoding the 4D scene into a novel spatio-temporal implicit neural map representation by fitting a time-dependent truncated signed distance function to each point.
  Using our representation, we extract the static map by filtering the dynamic parts.
  Our neural representation is based on sparse feature grids, a globally shared decoder, and time-dependent basis functions, which we jointly optimize in an unsupervised fashion.
  To learn this representation from a sequence of LiDAR scans, we design a simple yet efficient loss function to supervise the map optimization in a piecewise way.
  We evaluate our approach \footnote{Code: \url{https://github.com/PRBonn/4dNDF}} on various scenes containing moving objects in terms of the reconstruction quality of static maps and the segmentation of dynamic point clouds.
  The experimental results demonstrate that our method is capable of removing the dynamic part of the input point clouds while 
  reconstructing accurate and complete 3D maps, outperforming several state-of-the-art methods.
  \vspace{-0.35cm}
\end{abstract}

\section{Introduction}

Mapping using range sensors, like LiDAR or RGB-D cameras, is a fundamental task in computer vision and robotics. Often, we want to obtain accurate maps to support downstream tasks such as localization, planning, or navigation.
For achieving an accurate reconstruction of an outdoor environment, we have to account for dynamics caused by moving objects, such as vehicles or pedestrians.
Furthermore, dynamic object removal plays an important role in autonomous driving and robotics applications for creating digital twins for realistic simulation and high-definition mapping, where a static map is augmented with semantic and task-relevant information.

\begin{figure}
  \centering
  \begin{subfigure}{0.485\linewidth}
    \includegraphics[width=\linewidth]{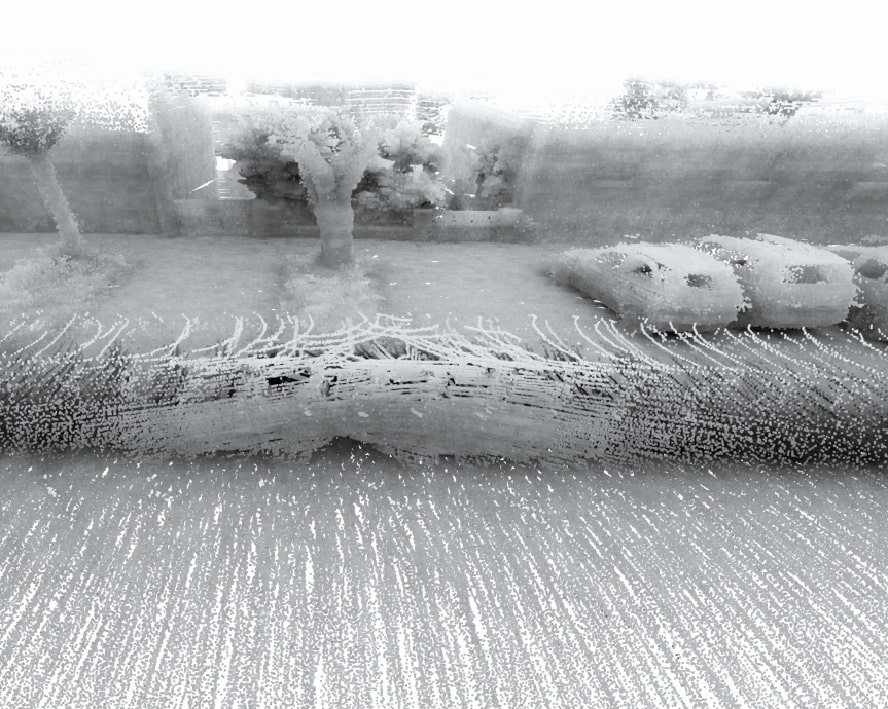}
    \subcaption{}
  \end{subfigure}
  \hfill
  \begin{subfigure}{0.485\linewidth}
    \includegraphics[width=\linewidth]{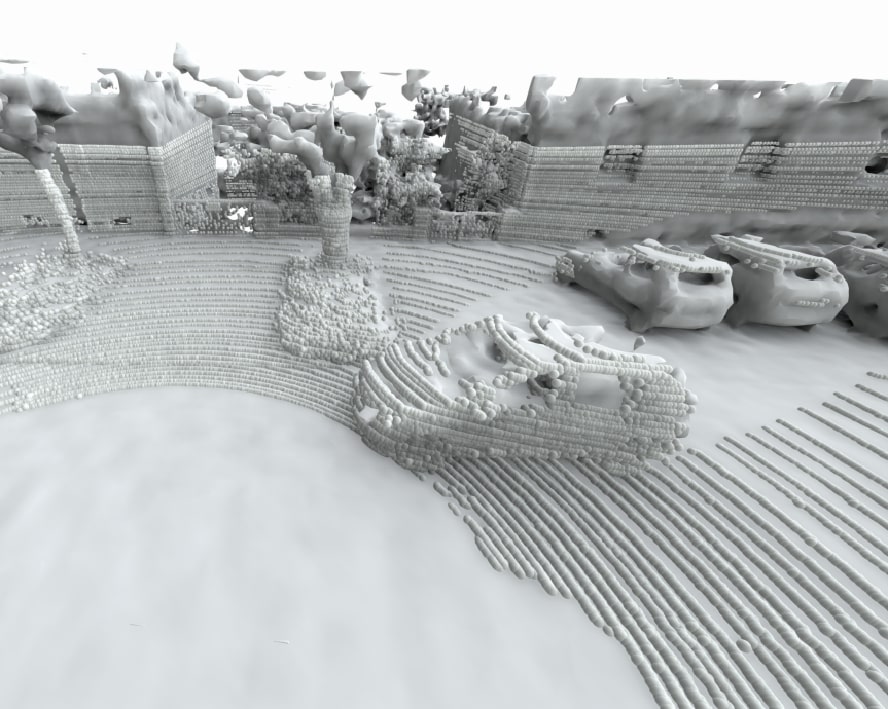}
    \subcaption{}
  \end{subfigure}
  \begin{subfigure}{0.485\linewidth}
    \includegraphics[width=\linewidth]{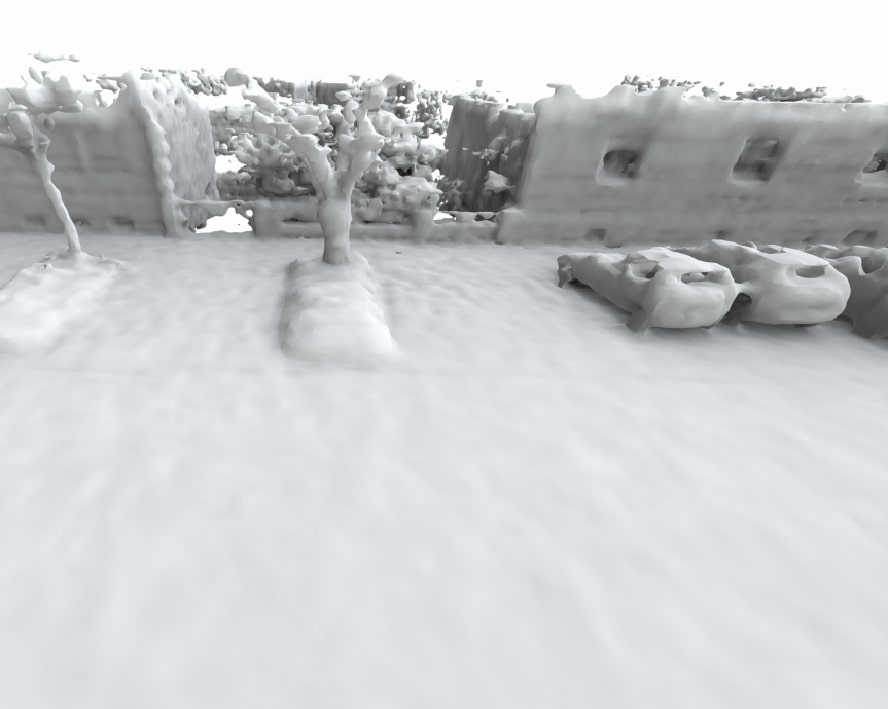}
    \subcaption{}
  \end{subfigure}
  \hfill
  \begin{subfigure}{0.485\linewidth}
    \includegraphics[width=\linewidth]{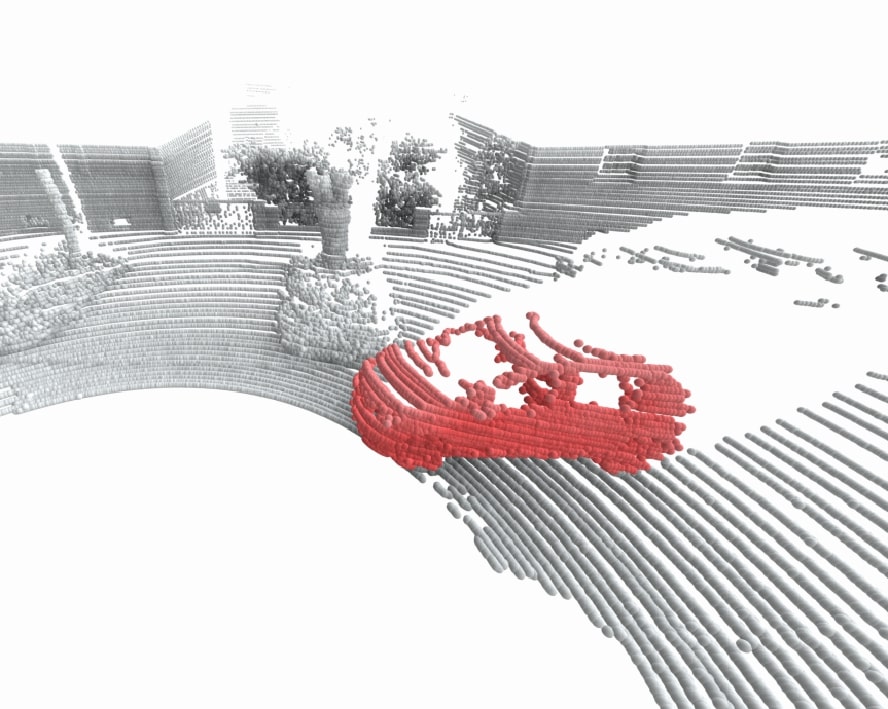}
    \subcaption{}
  \end{subfigure}
  \caption{Given a sequence of point clouds, as shown in (a), we optimize our 4D neural representation that can be queried at arbitrary positions for a specific time.
  (b) Based on the estimated time-dependent TSDF values, we can extract a mesh at a specific point in time. Additionally, our 4D neural representation can be also used for static mapping~(c) and dynamic object removal~(c). \vspace{-0.5cm}}
  \label{fig:motivation}
\end{figure}

Mapping and state estimation in dynamic environments is a classical problem in robotics~\cite{cadena2016tro, thrun2005probrobbook,stachniss2016handbook-slamchapter}.
Approaches for simultaneous localization and mapping~(SLAM) can apply different strategies to deal with dynamics. Common ways are:
(1) filtering dynamics from the input~\cite{barsan2018icra, runz2017icra, ruenz2018ismar, mccormac2017icra, salas-moreno2013cvpr} as a pre-processing step, which requires a semantic interpretation of the scene;
(2) modeling the occupancy in the map representation~\cite{newcombe2011ismar, saarinen2013iros-f3mi,saarinen2012iros,meyer-delius2012aaai,whelan2015rss,hornung2013ar}, where dynamics can be implicitly removed by retrospectively removing measurements in free space;
(3) including it in the state estimation~\cite{stachniss2005aaai, walcott-bryant2012iros,haehnel2002iros,biber2005rss,wolf2005ar} to model which measurements originated from the dynamic and static parts of the environment.
Our proposed method falls into the last category and allows us to model dynamics directly in the map representation leading to a spatio-temporal map representation.


Recently, implicit neural representations gained increasing interest in computer vision for novel view synthesis~\cite{mildenhall2020eccv, mueller2022acmgraphics} and 3D shape reconstruction~\cite{mescheder2019cvpr, park2019cvpr}.
Due to their compactness and continuity, several approaches~\cite{zhong2023icra, wiesmann2023ral, yan2023icra} investigate the use of neural representations in large-scale 3D LiDAR mapping leading to accurate maps while significantly reducing memory consumption.
However, these approaches often do not address the problem of handling dynamics during mapping. 
The recent progress on dynamic NeRF~\cite{sara2023cvpr, shao2023cvpr, cao2023cvpr, ramasinghe2023arxiv} and neural deformable object reconstruction~\cite{Cai2022neurips, chen2023pami} indicates that neural representations can be also used to represent dynamic scenes,
which inspires us to tackle the problem of mapping in dynamic environments from the perspective of 4D reconstruction.

In this paper, we propose a novel method to reconstruct large 4D dynamic scenes by encoding every point's time-dependent truncated signed distance function~(TSDF) into an implicit neural scene representation.
As illustrated in \cref{fig:motivation}, we take sequentially recorded LiDAR point clouds collected in dynamic environments as input and generate a TSDF for each time frame, which can be used to extract a mesh using marching cubes~\cite{lorensen1987siggraph}. The background TSDF, which is unchanged during the whole sequence, can
be extracted from the 4D signal easily. We regard it as a static map that can be used to segment dynamic objects from the original point cloud.
Compared to the traditional voxel-based mapping method, the continuous neural representation allows for the removal of dynamic objects while preserving rich map details.
In summary, the main contributions of this paper are:
\begin{itemize}
  \item We propose a novel implicit neural representation to jointly reconstruct a dynamic
        3D environment and maintain a static map using sequential LiDAR scans as input.
  \item We employ a piecewise training data sampling strategy and design a simple, yet effective loss function that maintains the consistency of the static point supervision through gradient constraints.
  \item We evaluate the mapping results by the accuracy of the dynamic object segmentation as well as the quality of the reconstructed static map showing superior performance compared to several baselines. We provide our code and the data used for experiments.
\end{itemize}

\section{Related Work}
\label{sec:related_work}

Mapping and SLAM in dynamic environments is a classical topic in robotics~\cite{cadena2016tro, thrun2005probrobbook,stachniss2016handbook-slamchapter} with a large body of work, which tackles the problem by pre-processing the sensor data~\cite{barsan2018icra, runz2017icra, ruenz2018ismar, mccormac2017icra, salas-moreno2013cvpr}, occupancy estimation to filter dynamics by removing measurements in free space~\cite{newcombe2011ismar, saarinen2013iros-f3mi,saarinen2012iros,meyer-delius2012aaai,whelan2015rss,hornung2013ar,palazzolo2019iros}, or state estimation techniques~\cite{stachniss2005aaai, walcott-bryant2012iros,haehnel2002iros,biber2005rss,wolf2005ar}.
Below, we focus on closely related approaches using neural representations but also static map building approaches for scenes containing dynamics.

\textbf{Dynamic NeRF.} Dynamic NeRFs aim to solve the problem of novel view synthesis in dynamic environments. Some approaches~\cite{park2021iccv, tretschk2021iccv, pumarola2021cvpr2, weng2022cvpr, park2021acm} address this challenge by
modeling the deformation of each point with respect to a canonical frame.
However, these methods cannot represent newly appearing objects. This can render them unsuited for complicated real-life scenarios.
In contrast, NSFF~\cite{li2021cvpr} and DynIBaR~\cite{li2023cvpr} get rid of the canonical frame by computing the motion field of the whole scene.
While these methods can deliver satisfactory results, the training time is usually in the order of hours or even days.

Another type of method leverages the compactness of the neural representation to model the 4D spatio-temporal information directly.
Several works~\cite{sara2023cvpr, cao2023cvpr, shao2023cvpr} project the 4D input into multiple voxelized lower-dimensional feature spaces to avoid large memory consumption,
which improves the efficiency of the optimization. Song \etal~\cite{song2023tvcg} propose a time-dependent sliding window strategy for accumulating the voxel features.
Instead of only targeting novel view synthesis, several approaches~\cite{yuan2021cvpr, Wu2022neurips, li2023cvpr} decompose the scene into dynamic objects and static background in a self-supervised way,
which inspired our work. Other approaches~\cite{Kundu2022cvpr, kong2023cvpr, song2023iccv} accomplish neural representation-based reconstruction for larger scenes by adding additional supervision such as object masks or optical flow.

\textbf{Neural representations for LiDAR scans.} Recently, many approaches aim to enhance scene reconstruction using LiDAR data through neural representations.
The early work URF~\cite{rematas2022cvpr} leverages LiDAR data as depth supervision to improve the optimization of a neural radiance field.
With only LiDAR data as input, Huang \etal~\cite{huang2023iccv} achieve novel view synthesis for LiDAR scans with differentiable rendering.
Similar to our work, Shine-mapping~\cite{zhong2023icra} and EINRUL~\cite{yan2023icra} utilize sparse hierarchical feature voxel structures to achieve large-scale 3D mapping.
Additionally, the data-driven approach NKSR~\cite{huang2023cvpr} based on learned kernel regression demonstrates accurate surface reconstruction with noisy LiDAR point cloud as input.
Although these approaches perform well in improving reconstruction accuracy and reducing memory consumption, none of them consider the problem of dynamic object interference in real-world environments.

\textbf{Static map building and motion detection.}
In addition to removing moving objects from the voxel map with ray tracing, numerous works~\cite{huang2022eccv,chen2021ral, mersch2022ral, mersch2023ral} try to segment dynamic points from raw LiDAR point clouds.
However, these methods require a significant amount of labeled data, which makes it challenging to generalize them to various scenarios or sensors with different scan patterns.
In contrast, geometry-based, more heuristic approaches have also produced promising results. Kim \etal~\cite{kim2020iros} solve this problem using the visibility of range images, but their results are still highly affected by the resolution.
Lim \etal proposed Erasor~\cite{lim2021ral}, which leverages ground fitting as prior to achieve better segmentation for dynamic points. More recent approaches~\cite{lim2023rss, chen2022ral} extend it to instance level to improve results.
However, these methods rely on an accurate ground fitting method, which is mainly designed for autonomous driving scenarios, which cannot be guaranteed in complex unstructured real environments.

In contrast to the approaches discussed above, we follow recent developments in neural reconstruction and propose a novel scene representation that allows us to capture the spatio-temporal progression of a scene.
We represent the time-varying SDF of a scene in an unsupervised fashion, which we exploit to remove dynamic objects and reconstruct accurate meshes of the static scene.
\section{Our Approach}
\label{sec:Method}

The input of our approach is given by a sequence of point clouds, $\mathcal{S}_{1:N} = (\mathcal{S}_1,\dots, \mathcal{S}_N)$, and their
corresponding global poses $\mq{T}_t \in \RR^{4\times4}$, $t \in [1, N]$, estimated via scan matching, LiDAR odometry, or SLAM methods~\cite{vizzo2023ral,behley2018rss,dellenbach2022icra,deschaud2018icra}.
Each scan's point cloud $\mathcal{S}_t = \{\vec{s}^1_t,\dots, \vec{s}^{M_t}_{t}\}$ is a set of points, $\vec{s}^i_t \in \RR^3$, collected at time $t$. 
Given such a sequence of scans $\mathcal{S}_{1:N}$, our approach aims to reconstruct a 4D TSDF of the traversed scene and maintain a static 3D map at the same time.

In the next sections, we first introduce our spatio-temporal representation and then explain how to optimize it to represent the dynamic and static parts of a point cloud sequence $\mathcal{S}_{1:N}$.

\subsection{Map Representation}
\label{subsec:maprep}
The key component of our approach is an implicit neural scene representation that allows us to represent a 4D TSDF of the scene, as well as facilitates the extraction of a static map representation.
Our proposed spatio-temporal scene representation is optimized for the given point cloud sequence $\mathcal{S}_{1:N}$ such that we can retrieve for an arbitrary point $\vec{p} \in \RR^3$ and time $t \in [1, N]$ the corresponding time-varying signed distances value at that location.


\textbf{Temporal representation. }
We utilize an TSDF to represent the scene, \ie, a function that provides the signed distance to the nearest surface for any given point $\vec{p} \in \RR^3$. The sign of the distance is positive when the point is in free space or in front of the measured surface and is negative when the point is inside the occupied space or behind the measured surface.

\begin{figure}[t]
  \centering
  \includegraphics[width=1.0\linewidth]{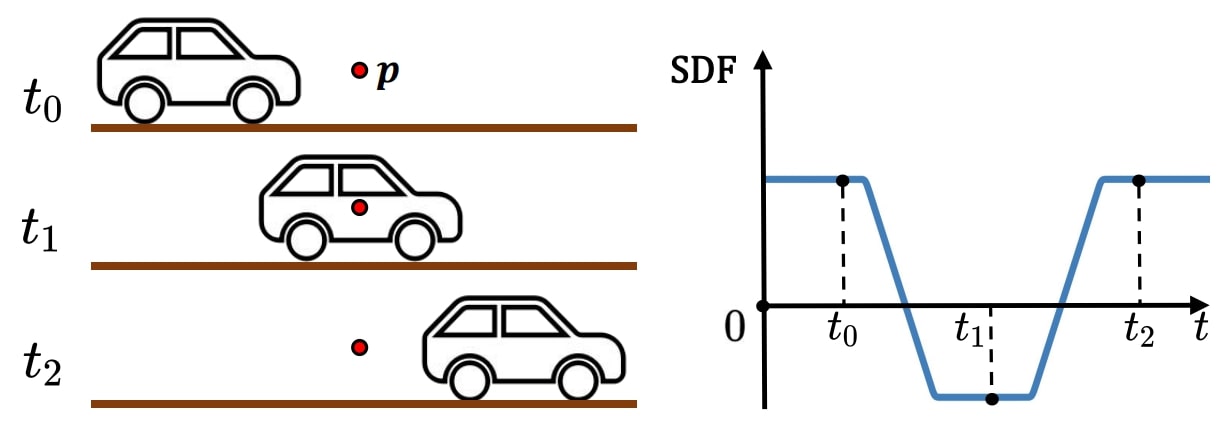}
  \caption{Principle of our 4D TSDF representation: The left figure shows a moving object and a query point $\vec{p}$. The one on the right depicts the corresponding signed distance at $\vec{p}$ over time. At $t_0$,  $\vec{p}$'s signed distance is a positive truncated value.
    When the moving object reaches $\vec{p}$ at time $t_1$, $\vec{p}$ is inside the object and its signed distance is negative accordingly.
    At $t_2$, the moving object moved past $\vec{p}$, the signed distance of $\vec{p}$ gets positive again.}
  \label{fig:curve}
  \vspace{-8pt}
\end{figure}

In a dynamic 3D scene, measuring the signed distance of any coordinate at each moment produces a time-dependent function
that captures the signed distance changes over time, see \cref{fig:curve} for an illustration. Additionally, if a coordinate is static
throughout the period, the signed distance should remain constant.
The key idea of our spatio-temporal scene representation is to fit the time-varying SDF at each point with several basis functions.
Inspired by Li \etal~\cite{li2023cvpr}'s representation of moving point trajectories,
we exploit $K$ globally shared basis functions $\phi_k: \RR \mapsto \RR$.
Using these basis functions~$\phi_k(t)$, we model the time-varying TSDF $F(\boldsymbol{p}, t)$ that maps a location $\vec{p}  \in \RR^3$ at time $t$ to a signed distance as follows:
\begin{align}
  F(\boldsymbol{p}, t) & = \sum_{k=1}^{K} w_{\boldsymbol{p} }^k\phi_k(t),
  \label{basis}
\end{align}
where $w_{\vec{p}}^k \in \RR$ are optimizable location-dependent coefficients. In line with previous works~\cite{li2023cvpr, wang2021neural}, we initialize the basis functions with discrete cosine transform~(DCT) basis functions:
\begin{align}
  \phi_k(t) & =\cos\left( \frac{\pi}{2N} (2t+1)(k-1)\right).
  \label{DCT}
\end{align}

The first basis function for $k=1$ is time-independent as $\phi_1(t) = 1$. During the training process, we fix $\phi_1(t)$ and determine the other basis functions by backpropagation.
We consider $\phi_1(t)$'s corresponding weight $w_{\boldsymbol{p} }^1$ as the static SDF value of the point $\boldsymbol{p}$.
Hence, $F(\vec{p}, t)$ consists of its static background value, \ie, $w^1_\vec{p}\phi_1(t) = w^1_\vec{p}$, and the weighted sum of dynamic basis functions $\phi_2(t), \dots, \phi_K(t)$.

As the basis functions $\phi_1(t), \dots, \phi_K(t)$ are shared between all points in the scene, we need to optimize the location-dependent weights that are implicitly represented in our spatial representation.

\begin{figure*}[t]
  \centering
  \includegraphics[width=1\linewidth]{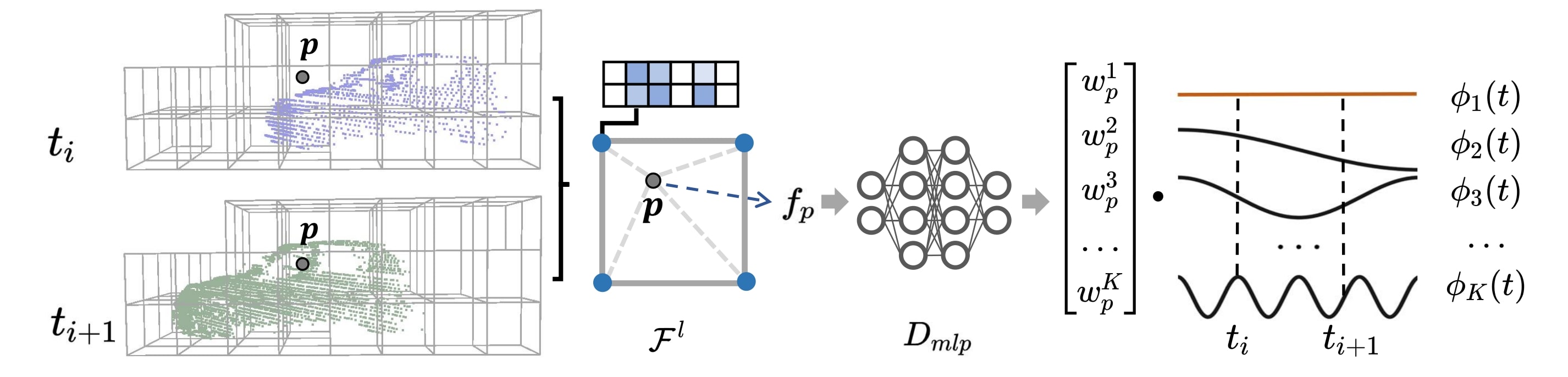}
  \caption{Overview of querying a TSDF value in our 4D map representation. For querying a point $\vec{p}$ at $t_{i}$ and $t_{i+1}$,
    we first retrieve each corner's feature in $\mathcal{F}^l$ of the voxel that $\vec{p}$ is located in and obtain the fused feature $\boldsymbol{f_{\boldsymbol{p}}}$ by trilinear interpolation.
    Then, we feed $\boldsymbol{f_{\boldsymbol{p}}}$ into the decoder $D_\text{mlp}$ and take the output as the weights of different basis functions $\phi_1(t), \dots, \phi_K(t)$.
    Finally, we calculate the weighted sum of basis functions' values at $t_{i}$ and $t_{i+1}$ to get their respective SDF results. For simplicity, we only illustrate one level of hashed feature grids.}
  \label{fig:overview}
  \vspace{-2pt}
\end{figure*}

\textbf{Spatial representation.}
To achieve accurate scene reconstruction while maintaining memory efficiency, we employ a multi-resolution
sparse voxel grid to store spatial geometric information.

First, we accumulate the input point clouds, $\mathcal{S}_1, \dots, \mathcal{S}_N$ based on their poses $\mq{T}_1, \dots, \mq{T}_N$ computed from LiDAR odometry and generate a hierarchy of voxel grids around points to ensure complete coverage in 3D.
We use a spatial hash table for fast retrieval of the resulting voxels that are only initialized if points fall into a voxel.

Similar to Instant-NGP~\cite{mueller2022acmgraphics},
we save a feature vector $\vec{f} \in \RR^D$ at each corner vertex of the voxel grid in each resolution level, where we denote as $\mathcal{F}^l$ the level-wise corner features.
We compute the feature vector $\vec{f}_\vec{p} \in \RR^D$ for given query point $\vec{p} \in \RR^3$ inside the hierarchical grid as follows:
\begin{align}
  \boldsymbol{f_{\boldsymbol{p}}} & = {\sum_{l=1}^{L} \text{interpolate}(\vec{p},\mathcal{F}^l)},
\end{align}
where $\text{interpolate}$ is the trilinear interpolation for a given point $\vec{p}$ using the corner features $\mathcal{F}^l$ at level $l$.

Then, we decode the interpolated feature vector $\vec{f}_\vec{p}$ into the desired weights $\vec{w}_\vec{p} = \left({w}^1_{p}, \dots, {w}^K_{p}\right) \in \RR^K$ by a globally shared multi-layer perceptron (MLP) $D_\text{mlp}$:
\begin{align}
  \boldsymbol{w}_{\boldsymbol{p}} & = D_\text{mlp}(\boldsymbol{\vec{f}_{\boldsymbol{p}}}).
\end{align}

As every step is differentiable, we can optimize the multi-resolution feature grids $\mathcal{F}^l$, the MLP decoder~$D_\text{mlp}$, and the values of the basis functions jointly by gradient descent once we have training data and corresponding target values.
The SDF querying process is illustrated in \cref{fig:overview}. 

\subsection{Objective Function}
\label{subsec:loss}

We take samples along the rays from the input scans $\mathcal{S}_t$ to collect training data.
Each scan frame $\mathcal{S}_t$ corresponds to a moment $t$ in time, so we gather four-dimensional data points $(\boldsymbol{q},t)$ via sampling along the ray from the scan origin \mbox{$\vec{o}_t \in \RR^3$} to a point $\vec{s}_t^i \in \mathcal{S}_t$.
We can represent the sampled points $\vec{q}_s^i$  along the ray as $\vec{q}_s^i = \vec{o}_t + \lambda(\vec{s}_t^i-\vec{o}_t)$. By setting a truncation threshold $\tau$, we split the ray into two regions, at the surface and in the free-space:
\begin{align}
  \mathcal{T}_\text{surf}^i & = {\left \{ \vec{q}_s^i \mid  \lambda \in \left( 1-\bar{\tau}, 1+\bar{\tau} \right) \right \}  } \\
  \mathcal{T}_\text{free}^i & = {\left \{ \vec{q}_s^i \mid  \lambda \in \left( 0,1-\bar{\tau} \right) \right \}  },
  \label{split_sampling}
\end{align}
where $\bar{\tau} = \tau \, (\norm{\vec{s}_t^i-\vec{o}_t})^{-1}$.
Thus, $\mathcal{T}_\text{surf}^i$~represents the region close to the endpoint $\vec{s}_t^i \in \mathcal{S}_t$, and $\mathcal{T}_\text{free}^i$ is the region in the free space.
We uniformly sample $M_s$ and $M_f$ points from $\mathcal{T}_\text{surf}^i$ and $\mathcal{T}_\text{free}^i$ separately. 
We obtain two sets $\mathcal{D}_\text{surf}$ and $\mathcal{D}_\text{free}$ of samples by sampling over all scans.
Unlike prior work~\cite{rematas2022cvpr, huang2023iccv} that use differentiable rendering to calculate the depth by integration along the ray, we design different losses for $\mathcal{D}_\text{surf}$ and $\mathcal{D}_\text{free}$ to supervise the 4D TSDF directly.

\textbf{Near Surface Loss.} Since the output of our 4D map is the signed distance value $\hat{d} = F(\vec{p}, t)$ at an arbitrary position $\vec{p} \in \RR^3$ in time $t \in [1, N]$, we expect that the predicted value $\hat{d}$ does not change over time for static points.
However, this cannot be guaranteed if we use the projective distance $d_\text{surf}$ to the surface along the ray direction directly as the target value, since the projective distance would change over time due to the change of view direction by the moving sensor, even in a static scene.
Thus, for the sampled data in $\mathcal{D}_\text{surf}$, \ie, the sampled points near the surface, we can only obtain reliable information about the sign of the TSDF value of these points, which should be positive if the point is before the endpoint and negative if the point is behind.
In addition, for a sampled point in front of the endpoint, its projective signed distance $d_\text{surf}$ should be the upper bound of its actual signed distance value. And for sampled points behind the endpoint,
$d_\text{surf}$ should be the lower bound.

We design a piecewise loss $L_\text{surf}$ to supervise the sampled points near the surface:
\begin{equation}
  L_\text{surf}(\hat{d}, d_\text{surf}) =\left\{
  \begin{array}{cl}
    |\,\hat{d}\,|               & \text{if } \hat{d} \, d_\text{surf} < 0               \\
    |\,\hat{d}-d_\text{surf}\,| & \text{if } \hat{d} \, d_\text{surf} > d_\text{surf}^2 \\
    0                           & \text{otherwise}                                        \\
  \end{array}
  \right.,
\end{equation}
where $\hat{d} = F(\boldsymbol{q}, t)$ is the predicted value from our map for a sample point $\boldsymbol{q} \in \mathcal{D}_\text{surf}$ and $d_\text{surf}$ is its corresponding projective signed distance for that sampled point in the corresponding scan $\mathcal{S}_t$.
This loss punishes only a prediction when the sign is wrong or its absolute value is larger than the absolute value of $d_\text{surf}$.
For a query point exactly on the surface, \ie,  $d_\text{surf} = 0$, $L_\text{surf}$ is simply the L1 loss.

To calculate an accurate signed distance value and maintain the consistency of constraints for static points from different observations, we use the natural property of signed distance function to constraint the length of the gradient vector for samples inside $\mathcal{D}_\text{surf}$, which is called Eikonal regularization ~\cite{ortiz2022rss, Gropp2020icml}:
\begin{equation}
  L_\text{eikonal} (\boldsymbol{p}, t) = \left (  \left \| \frac{\partial F(\boldsymbol{p}, t) }{\partial \boldsymbol{p}} \right \|   -1 \right )^{2}  ,
  \label{batch_loss}
\end{equation}

Inspired by Neuralangelo~\cite{lizhao2023cvpr}, we manually add perturbations to compute more robust gradient vectors instead of using automatic differentiation, which means we compute numerical gradients:
\begin{equation}
  \label{numrical gradient}
  \nabla_x F(\boldsymbol{p}, t) = \frac{F(\boldsymbol{p}+\boldsymbol{\epsilon_x}, t)-F(\boldsymbol{p}-\boldsymbol{\epsilon_x}, t)}{2\epsilon},
\end{equation}
where $\nabla_x F(\boldsymbol{p}, t)$ is the component of the gradient $\frac{\partial F(\boldsymbol{p}, t)}{\partial \boldsymbol{p}}$ on the $x$ axis, and $\boldsymbol{\epsilon_x} = (\epsilon,0,0)^\top$ is the added perturbation.
We apply the same operation on $y$ and $z$ axes to calculate the numerical gradient.
Furthermore, in order to get faster convergence at the beginning and ultimately recover the rich geometric details, we first set a large $\epsilon$ and gradually reduce it during the training process. 

\textbf{Free Space Loss.}
As we tackle the problem of mapping in dynamic environments, we cannot simply accumulate point clouds and then calculate accurate supervision of signed distance value via nearest neighbor search.
Therefore, we use a L1 loss $L_\text{free}$ to constrain the signed distance prediction $\hat{d}$ of the free space points, \ie, $\vec{p} \in \mathcal{D}_\text{free}$:
\begin{equation}
  L_\text{free}(\hat{d}) = \lvert \hat{d} - \tau \rvert,
  \label{free_loss}
\end{equation}
where $\tau$ is the truncation threshold we used in \cref{subsec:loss}. 


Thanks to our spatio-temporal representation, a single query point can get both, static and dynamic TSDF values.
Thus, for some regions that are determined to be free space, we can directly add constraints to their static TSDF values.

We divide the free space points $\mathcal{D}_\text{free}$ into dense and sparse subset $\mathcal{D}_\text{dense}$ and $\mathcal{D}_\text{sparse}$ based on a threshold $r_\text{dense}$ for the distance from the free space point sampled at time $t$ to the scan origin $\vec{o}_t$. For each point $\vec{p} \in \mathcal{D}_\text{dense}$, we find the nearest neighbor $\vec{n}_\vec{p}$ in the corresponding scan $\mathcal{S}_t$, \ie, $\vec{n}_\vec{p} = \arg\min_{\vec{q}\in \mathcal{S}_t} \norm{\vec{p}- \vec{q}}_2$.
Let $\mathcal{D}_\text{certain} = \{ \vec{p} \in \mathcal{D}_\text{dense} \mid ||\vec{p} - \vec{n}_\vec{p}|| > \tau \}$ be the points that we consider in the certain free space. Then, we supervise $\vec{p} \in \mathcal{D}_\text{certain}$ by its static signed distance value directly:
\begin{align}
  L_\text{certain}(\vec{p}) & = \lvert w_\vec{p}^1- \tau  \rvert,
  \label{certain free_loss}
\end{align}
where $w_\vec{p}^1$ is the first weight of the decoder's output.

\begin{figure}[!t]
  \centering
  \begin{subfigure}[b]{0.195\textwidth}
    \centering
    \includegraphics[height=0.92in]{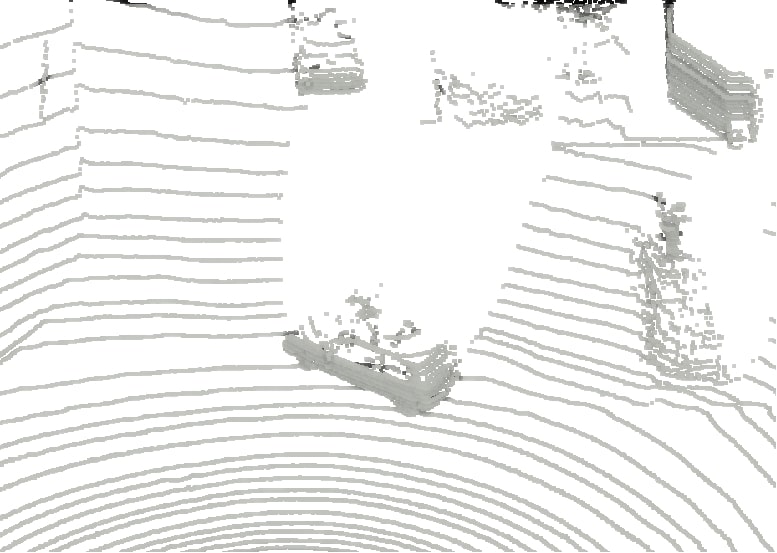}
    \subcaption{}
    \label{subfig:(a)}
  \end{subfigure}
  \hfil
  \begin{subfigure}[b]{0.195\textwidth}
    \centering
    \includegraphics[height=0.92in]{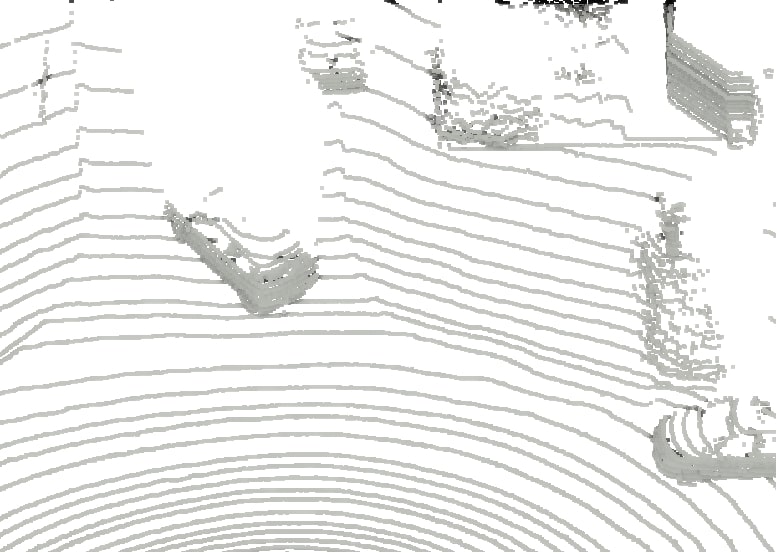}
    \subcaption{}
    \label{subfig:(b)}
  \end{subfigure}

  \begin{subfigure}[b]{0.195\textwidth}
    \centering
    \includegraphics[height=0.92in]{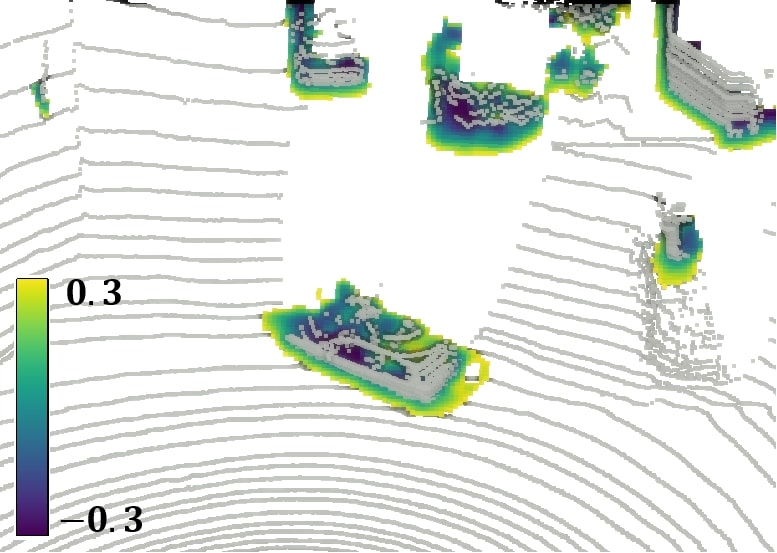}
    \subcaption{}
    \label{subfig:(c)}
  \end{subfigure}
  \hfil
  \begin{subfigure}[b]{0.195\textwidth}
    \centering
    \includegraphics[height=0.92in]{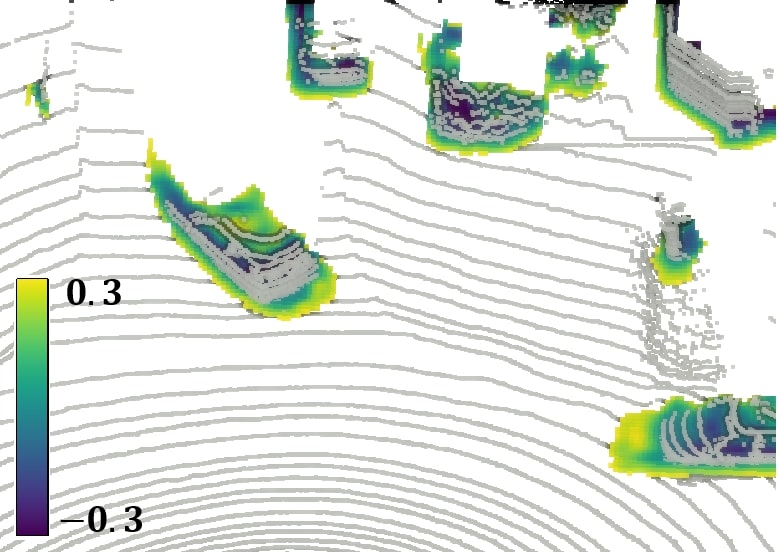}
    \subcaption{}
    \label{subfig:(d)}
  \end{subfigure}
  \vspace{-4pt}
  \caption{Reconstructed TSDF for KITTI dataset~\cite{geiger2012cvpr}: Subfigures (a) and (b) are the input neighboring frames.
    Correspondingly, (c) and (d) are horizontal TSDF slices queried from our 4D map.
    Note that we only display the TSDF values that are less than 0.3m.}
  \label{fig:sdf}
  \vspace{-8pt}
\end{figure}

In summary, the final loss $L_\text{total}$ is given by:
\begin{align}
  L_\text{total} & = \frac{1}{|\mathcal{D}_\text{surf}|}\sum_{(\boldsymbol{p},t)\in\mathcal{D}_{\text{surf}}} L_\text{surf}(\hat{d}, d_\text{surf}) + \lambda_{e} L_\text{eikonal}(\vec{p}, t)\nonumber \\
                 & + \frac{\lambda_{f}}{|\mathcal{D}_\text{free}|}\sum_{(\boldsymbol{p},t)\in\mathcal{D}_{\text{free}}}  L_\text{free}(\hat{d}) \nonumber                                               \\
                 & + \frac{\lambda_{c}}{|\mathcal{D}_\text{certain}|}\sum_{(\boldsymbol{p},t)\in\mathcal{D}_{\text{certain}}}   L_\text{certain}(\vec{p}),
  \label{total loss}
\end{align}
where $\hat{d} = F(\vec{p}, t)$ is the predicted signed distance at the sample position $\vec{p}$ at time $t$ and $d_\text{surf}$ is the projective signed distance of sample $\vec{p}$.
With the above loss function and data sampling strategy, we train our map offline until convergence. 
In \cref{fig:sdf}, we show TSDF slices obtained using our optimized 4D map at different times.

One application of our 4D map representation is dynamic object segmentation. For a point $\vec{p}$ in the input scans $\mathcal{S}_{1:N}$, 
its static signed distance value $w_\vec{p}^1$ can be obtained by a simple query.
If $\vec{p}$ belongs to the static background, it should have $w_\vec{p}^1 = 0$. Therefore, we simply set a threshold $d_\text{static}\,$ and regard a point as dynamic if $w_\vec{p}^1 > d_\text{static}$.

\subsection{Implementation Details}
\label{subsec:Implementation}
As hyperparameters of our approach, we use the values listed in \cref{tab:implementation} in all LiDAR experiments.
Additional parameters are determined by the characteristics of the sensor and the dimensions of the scene.
For instance, in the reconstruction of autonomous driving scenes, like KITTI,
we set the highest resolution for the feature voxels to $0.3$\,m. The truncation distance is set to $\tau = 0.5$\,m, 
and the dense area split threshold $r_\text{dense} = 15\,$m.
Regarding training time, it takes $12$ minutes to train $140$ frames from the KITTI dataset using a single Nvidia Quadro RTX 5000.

\section{Experiments}
\label{sec:experiments}
In this section, we show the effectiveness of our proposed approach with respect to two aspects:
(1) Static mapping quality: The static TSDF built by our method allows us to extract a surface mesh using marching cubes~\cite{lorensen1987siggraph}.
We compare this extracted mesh with the ground truth mesh to evaluate the reconstruction.
(2) Dynamic object segmentation: As mentioned above, our method can segment out the dynamic objects in the input scans. 
We use point-wise dynamic object segmentation accuracy to evaluate the results.

\subsection{Static Mapping Quality}
\label{subsec:staticMQ}

\textbf{Datasets.} We select two datasets collected in dynamic environments for quantitative evaluation.
One is the synthetic dataset \textit{ToyCar3} from Co-Fusion~\cite{runz2017icra},
which provides accurate depth images and accurate masks of dynamic objects rendered using Blender, but also depth images with added noise.
For this experiment, we select 150 frames from the whole sequence, mask out all dynamic objects in the accurate depth images, 
and accumulate background static points as the ground-truth static map.  The original noisy depth images are used as the input for all methods.

Furthermore, we use the \textit{Newer College}~\cite{ramezani2020iros} dataset as the real-world dataset, which is collected using a 64-beam LiDAR.
Compared with synthetic datasets, it contains more uncertainty from measurements and pose estimates.
We select 1,300 frames from the courtyard part for testing and this data includes a few pedestrians as dynamic objects.
This dataset offers point clouds obtained by a high-precision terrestrial laser scanner that can be directly utilized as ground truth to evaluate the mapping quality.

\textbf{Metric and Baselines.}
We report the reconstruction accuracy, completeness, the Chamfer distance, and the F1-score.
Further details on the computation of the metrics can be found in the supplement.

We compare our method with several different types of state-of-the-art methods:
(i) the traditional TSDF-fusion method, VDBfusion~\cite{vizzo2022sensors},
which uses space carving to eliminate the effects of dynamic objects,
(ii) the data-driven-based method, neural kernel surface reconstruction~(NKSR)~\cite{huang2023cvpr}, and
(iii) the neural representation based 3D mapping approach, SHINE-mapping~\cite{zhong2023icra}.

For NKSR~\cite{huang2023cvpr}, we use the default parameters provided by Huang \etal with their official implementation.
To ensure a fair comparison with SHINE-mapping, we adopt an equal number of free space samples (15 samples), aligning with our method for consistency.

For the \textit{ToyCar3} dataset, we set VDB-Fusion's resolution to $1\,$cm. To have all methods with a similar
memory consumption, we set the resolution of SHINE-mapping's leaf feature voxel to $2$\,cm, and our method's highest resolution accordingly to $2\,$cm.
For the \textit{Newer College} dataset, we set the resolution to $10$\,cm, $30$\,cm, and $30$\,cm respectively.

\begin{table}[t]
  \centering
  \caption{Hyperparameters of our approach.}
  \setlength{\belowcaptionskip}{-3pt}
  \small{
  \centering
  \setlength{\tabcolsep}{1.1mm}{
    \begin{tabular}{lcccccccc}
      \toprule
      \textbf{Parameter} & \textbf{Value} & \textbf{Description}               \\ \midrule
      L                  & 2              & number of feature voxels level     \\
      D                  & 8              & The length of feature vectors      \\
      K                  & 32             & The number of basis functions      \\
      $D_\text{mlp}$     & $2 \times 64$  & layer and size of the MLP decoder                      \\\midrule
      $M_s$              & 5              & The number of surface area samples \\
      $M_f$              & 15             & The number of free space samples   \\\midrule
      $\lambda_{e}$      & 0.02            & weight for Eikonal loss            \\
      $\lambda_{f}$      & 0.25            & weight for free space loss         \\
      $\lambda_{c}$      & 0.2            & weight for certain free loss       \\ \bottomrule
    \end{tabular}
  }
  }
  \label{tab:implementation}
  \vspace{-2pt}
\end{table}

\begin{figure*}[ht]
  \centering
  \begin{subfigure}[b]{0.195\textwidth}
    \centering
    \includegraphics[height=1.0in]{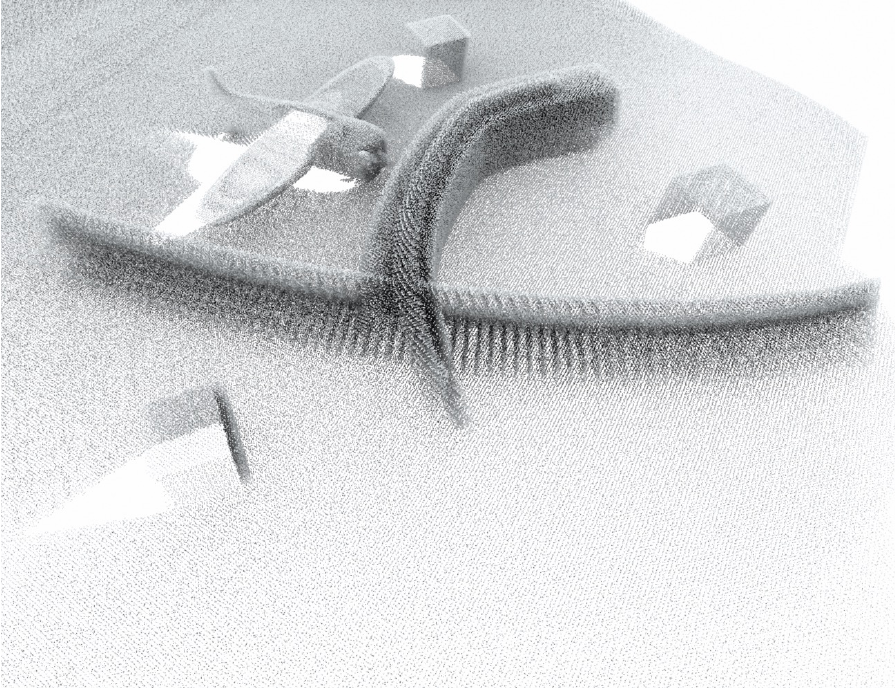}
    \label{subfig:toycar_pc}
    \subcaption{Merged input scans}
  \end{subfigure}
  \hfill
  \begin{subfigure}[b]{0.195\textwidth}
    \centering
    \includegraphics[height=1.0in]{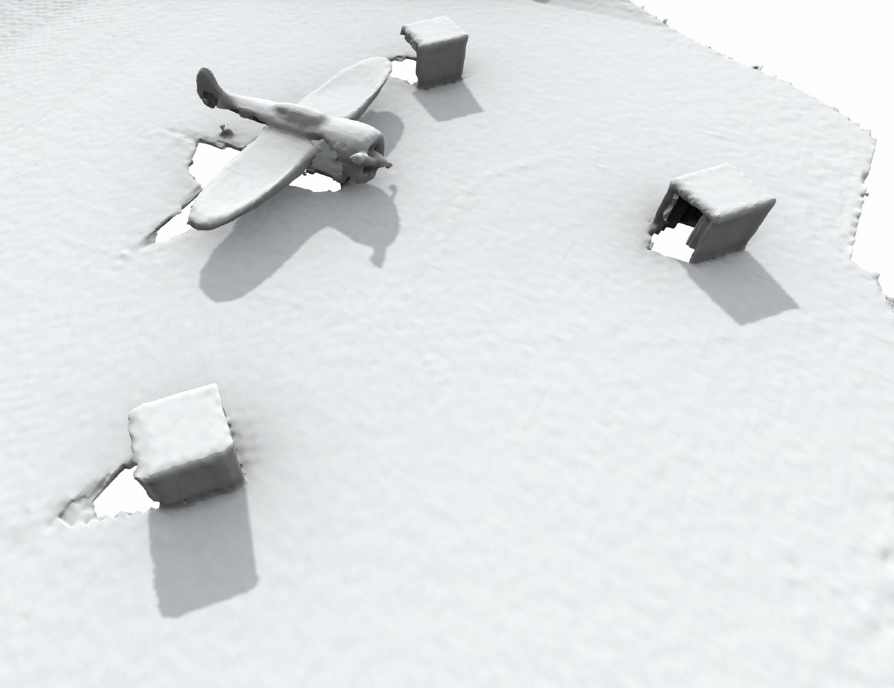}
    \label{subfig:toycar_ours}
    \subcaption{Ours}
  \end{subfigure}
  \hfill
  \begin{subfigure}[b]{0.195\textwidth}
    \centering
    \includegraphics[height=1.0in]{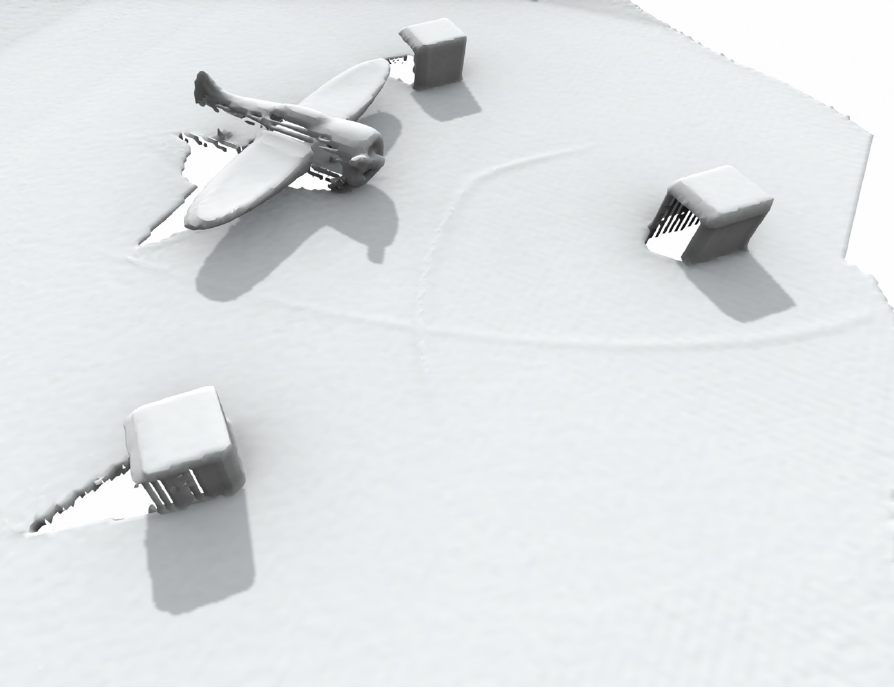}
    \label{subfig:toycar_vdb}
    \subcaption{VDB-Fusion~\cite{vizzo2022sensors}}
  \end{subfigure}
  \hfill
  \begin{subfigure}[b]{0.195\textwidth}
    \centering
    \includegraphics[height=1.0in]{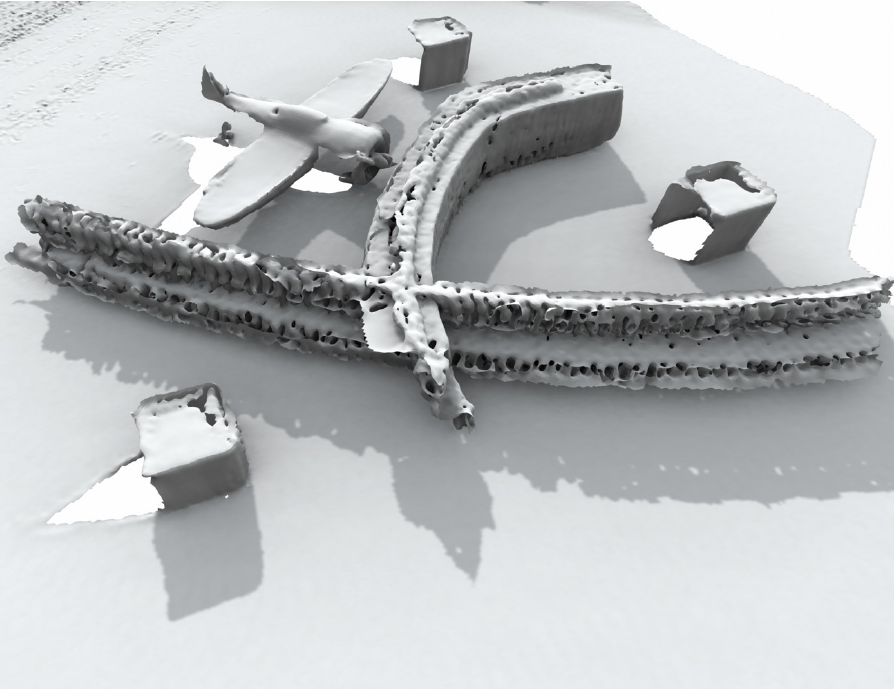}
    \label{subfig:toycar_nksr}
    \subcaption{NKSR~\cite{huang2023cvpr}}
  \end{subfigure}
  \hfill
  \begin{subfigure}[b]{0.195\textwidth}
    \centering
    \includegraphics[height=1.0in]{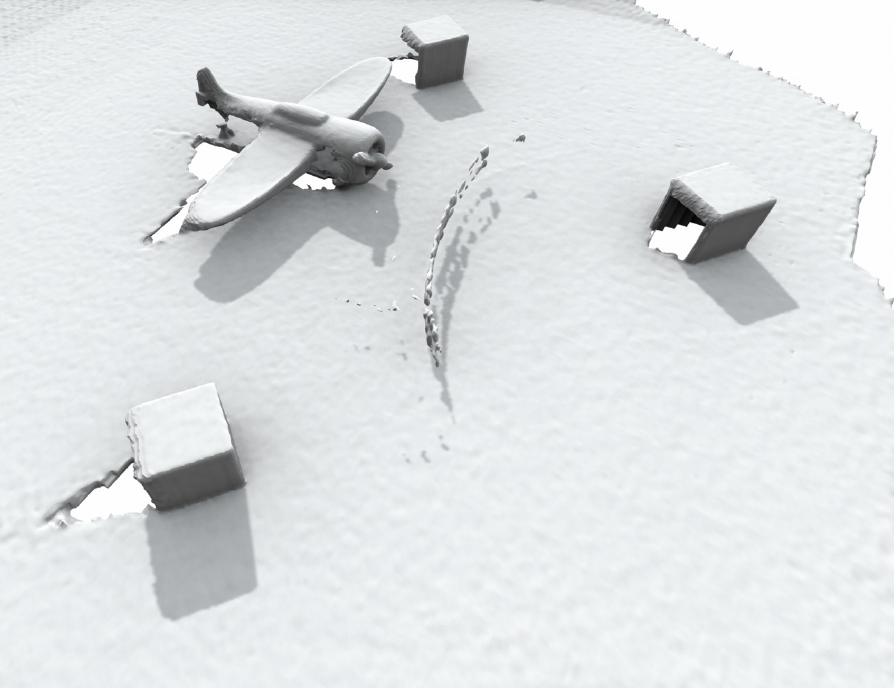}
    \label{subfig:toycar_shine}
    \subcaption{SHINE-mapping~\cite{zhong2023icra}}
  \end{subfigure}
  \vspace{-4pt}
  \caption{A comparison of the static mapping results of different methods on the  \textit{ToyCar3} dataset. There are two dynamic toy cars moving through the scene. Our method can reconstruct the static scene with fine details and eliminate the dynamic car.}
  \label{fig:toycar_recon_exp}
  \vspace{-4pt}
\end{figure*}

\begin{figure*}[ht]
  \centering
  \begin{subfigure}[b]{0.195\textwidth}
    \centering
    \includegraphics[height=1.75in]{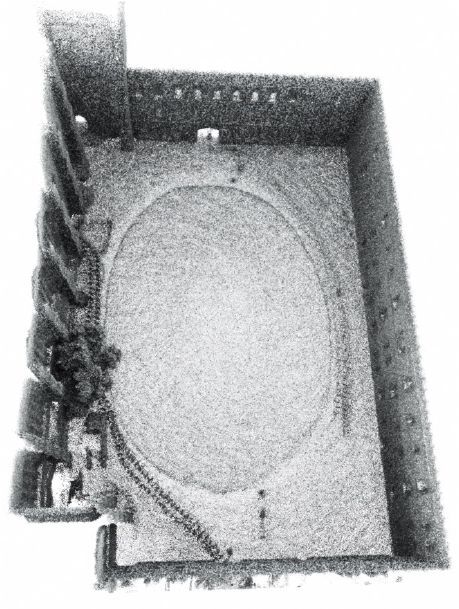}
    \label{subfig:ncd_pc}
    \vspace{-12pt}
    \subcaption{Merged input scans}
  \end{subfigure}
  \hfill
  \begin{subfigure}[b]{0.195\textwidth}
    \centering
    \includegraphics[height=1.75in]{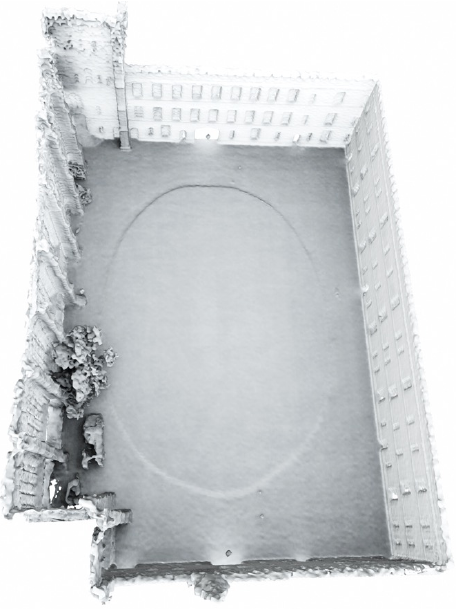}
    \label{subfig:ncd_ours}
    \vspace{-12pt}
    \subcaption{Ours}
  \end{subfigure}
  \hfill
  \begin{subfigure}[b]{0.195\textwidth}
    \centering
    \includegraphics[height=1.75in]{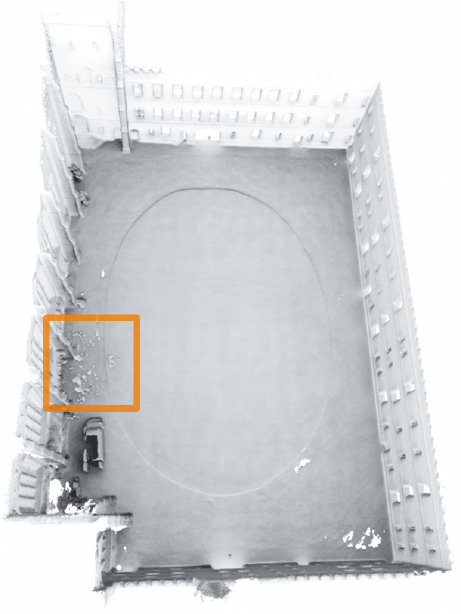}
    \label{subfig:ncd_vdb}
    \vspace{-12pt}
    \subcaption{VDB-Fusion~\cite{vizzo2022sensors}}
  \end{subfigure}
  \hfill
  \begin{subfigure}[b]{0.195\textwidth}
    \centering
    \includegraphics[height=1.75in]{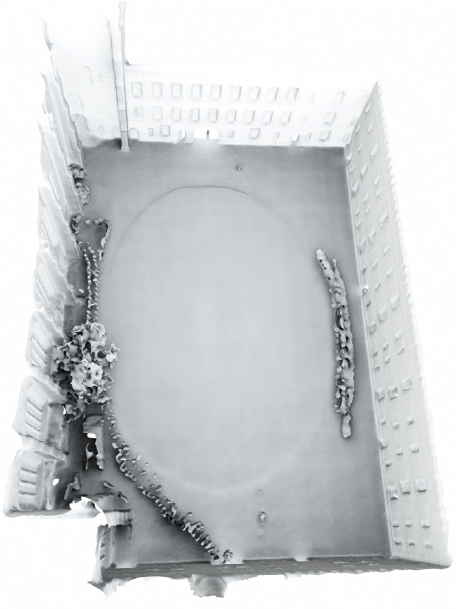}
    \label{subfig:ncd_nksr}
    \vspace{-12pt}
    \subcaption{NKSR~\cite{huang2023cvpr}}
  \end{subfigure}
  \hfill
  \begin{subfigure}[b]{0.195\textwidth}
    \centering
    \includegraphics[height=1.75in]{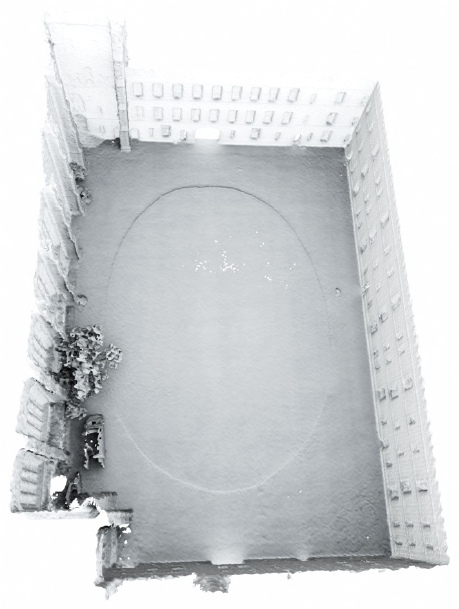}
    \label{subfig:ncd_shine}
    \vspace{-12pt}
    \subcaption{SHINE-mapping~\cite{zhong2023icra}}
  \end{subfigure}
  \vspace{-4pt}
  \caption{A comparison of the static mapping results of different methods on the \textit{Newer College} dataset. Several pedestrians are moving through the scene during the data collection. 
  Our method can reconstruct the static scene completely and eliminate the moving pedestrians. 
  Although VDB-Fusion manages to eliminate the pedestrians, it incorrectly removes the tree highlighted in the orange box.}
  \label{fig:ncd_recon_exp}
  \vspace{-8pt}
\end{figure*}

\textbf{Results.}
The quantitative results for synthetic dataset \textit{ToyCar3} and real-world dataset \textit{Newer College} are presented in \cref{tab:experiments_on_ToyCar3} and \cref{tab:experiments_on_NewerCollege}, respectively. 
We also show the extracted meshes from all methods in \cref{fig:toycar_recon_exp} and \cref{fig:ncd_recon_exp}.

Our method outperforms the baselines in terms of Completeness and Chamfer distance for both datasets (\cf \cref{fig:toycar_recon_exp} and \cref{fig:ncd_recon_exp}).
Regarding the accuracy, SHINE-mapping and VDB-Fusion can filter part of high-frequency noise by fusion of multiple frames, resulting in better performance on noisy \textit{Newer College} dataset. 
In comparison, our method considers every scan as accurate to store 4D information, which makes it more sensitive to measurement noise.
On the \textit{ToyCar3} dataset, both our method and VDB-Fusion successfully
eliminate all moving objects. However, on the \textit{Newer College} dataset, VDB-Fusion incorrectly eliminates the static tree
and parts of the ground, resulting in poor completeness shown in \cref{tab:experiments_on_NewerCollege}.
SHINE-mapping eliminates dynamic pedestrians on the \textit{Newer College} dataset 
but retains a portion of the dynamic point cloud on the \textit{ToyCar3} dataset, which has a larger proportion of dynamic objects,
leading to poorer accuracy in \cref{tab:experiments_on_ToyCar3}.
NKSR performs the worst accuracy because it is unable to eliminate dynamic objects, which means it's not suitable to apply NKSR
in dynamic real-world scenes directly.

\begin{table}[t]
  \caption{Quantitative results of the reconstruction quality on \textit{ToyCar3}.
    We report the distance error metrics, namely completion, accuracy and Chamfer-L1 in cm.
    Additionally, we show the F-score in $\%$ with a $1$\,cm error threshold.}
  \setlength{\belowcaptionskip}{-2pt}
  \centering
  \small{
  \setlength{\tabcolsep}{1mm}{
    \begin{tabular}{lccccc}
      \toprule
      \textbf{Method} & \textbf{Comp.~$\downarrow$} & \textbf{Acc.~$\downarrow$} & \textbf{C-L1~$\downarrow$} & \textbf{F-score~$\uparrow$} \\ \midrule
      VDB-fusion~\cite{vizzo2022sensors}      & 0.574                       & 0.481             & 0.528                      & 97.95                       \\
      NKSR~\cite{huang2023cvpr}            & 0.526                       & 2.809                      & 1.667                      & 89.54                    \\
      SHINE-mapping~\cite{zhong2023icra}           & 0.583                       & 0.626                      & 0.605                      & 98.01                       \\ \midrule
      Ours            & \textbf{0.438}              & \textbf{0.468}                      & \textbf{0.452}             & \textbf{98.35}               \\ \bottomrule
    \end{tabular}
  }}
  \label{tab:experiments_on_ToyCar3}
  \vspace{-5pt}
\end{table}

\begin{table}[t]
  \caption{Quantitative results of the reconstruction quality on \textit{Newer College}.
    We report the distance error metrics, namely completion, accuracy and Chamfer-L1 in cm.
    Additionally, we show the F-score in $\%$ with a 20\,cm error threshold.}
  \setlength{\belowcaptionskip}{-3pt}
  \centering
  \small{
  \setlength{\tabcolsep}{1mm}{
    \begin{tabular}{lcccc}
      \toprule
      \textbf{Method} & \textbf{Comp.~$\downarrow$} & \textbf{Acc.~$\downarrow$} & \textbf{C-L1~$\downarrow$} & \textbf{F-score~$\uparrow$} \\ \midrule
      VDB-fusion~\cite{vizzo2022sensors}      & 7.32                        & 5.99                       & 6.65                       & 96.68                       \\
      NKSR~\cite{huang2023cvpr}            & 6.87                        & 9.28                       & 8.08                       & 95.65                       \\
      SHINE-mapping~\cite{zhong2023icra}           & 6.80                        & \textbf{5.86}              & 6.33                       & \textbf{97.67}              \\ \midrule
      Ours            & \textbf{5.85}               & 6.49                       & \textbf{6.17}             & 97.50                       \\ \bottomrule
    \end{tabular}
  }
  }
  \label{tab:experiments_on_NewerCollege}
  \vspace{-0.3cm}
\end{table}

\begin{table*}[t]
  \caption{Quantitative results of the dynamic object removal quality on the KTH-Dynamic-Benchmark.
    We report the static accuracy SA, dynamic static DA and the associated accuracy AA.
    Octomap* refers to the modified Octomap implementation by Zhang \etal~\cite{zhang2023ITSC}.}
  \setlength{\belowcaptionskip}{-8pt}
  \centering
  \small{
  \setlength{\tabcolsep}{2.6mm}{
    \begin{tabular}{lccccccccccccc}
      \toprule
               & \multicolumn{3}{c}{KITTI Seq. 00} & \multicolumn{3}{c}{KITTI Seq. 05} & \multicolumn{3}{c}{Argoverse2} & \multicolumn{3}{c}{Semi-Indoor}                                                                                                                          \\
      \cmidrule(lr){2-4}\cmidrule(lr){5-7}\cmidrule(lr){8-10}\cmidrule(lr){11-13}
      Method   & \textbf{SA}                       & \textbf{DA}                       & \textbf{AA}                    & \textbf{SA}                     & \textbf{DA} & \textbf{AA}    & \textbf{SA} & \textbf{DA} & \textbf{AA}    & \textbf{SA} & \textbf{DA} & \textbf{AA}    \\ \midrule
      Octomap~\cite{hornung2013ar}  & 68.05                             & \textbf{99.69}                             & 82.37                          & 66.28                           & \textbf{99.24}       & 81.10          & 65.91       & 96.70       & 79.84          & 88.97       & \textbf{82.18}       & \textbf{85.51} \\
      Octomap*~\cite{zhang2023ITSC} & 93.06                             & 98.67                             & 95.83                          & 93.54                           & 92.48       & 93.01          & 82.66       & 82.44       & 82.55          & 96.79       & 73.50       & 84.34          \\
      Removert~\cite{kim2020iros} & 99.44                             & 41.53                             & 64.26                          & 99.42                           & 22.28       & 47.06          & 98.97       & 31.16       & 55.53          & 99.96       & 12.15       & 34.85          \\
      Erasor~\cite{lim2021ral}   & 66.70                              & 98.54                             & 81.07                          & 69.40                            & 99.06       & 82.92          & 77.51       & \textbf{99.18}       & 87.68          & 94.90        & 66.26       & 79.30           \\
      SHINE~\cite{zhong2023icra}    & 98.99                             & 92.37                             & 95.63                          & 98.91                           & 53.27       & 72.58          & 97.66       & 72.62       & 84.21          & 98.88       & 59.19       & 76.51          \\
      4DMOS~\cite{mersch2022ral}    & -                                 & -                                 & -                              & -                               & -           & -              & 99.94       & 69.33       & 83.24          & \textbf{99.99}       & 10.60       & 32.55          \\
      MapMOS~\cite{mersch2023ral}   & -                                 & -                                 & -                              & -                               & -           & -              & \textbf{99.96}       & 85.88       & 92.65          & \textbf{99.99}       & 4.75        & 21.80          \\ \midrule
      Ours     & \textbf{99.46}                             & 98.47                             & \textbf{98.97}                 & \textbf{99.54}                           & 98.36       & \textbf{98.95} & 99.17       & 95.91       & \textbf{97.53} & 94.17      & 72.79       & 82.79       \\ \bottomrule
    \end{tabular}
  }}
  \label{tab:experiments_on_KTH}
  \vspace{-2pt}
\end{table*}

\begin{figure*}[t]
  \centering
  \begin{subfigure}{0.195\textwidth}
    \centering
    \includegraphics[height=1.1in]{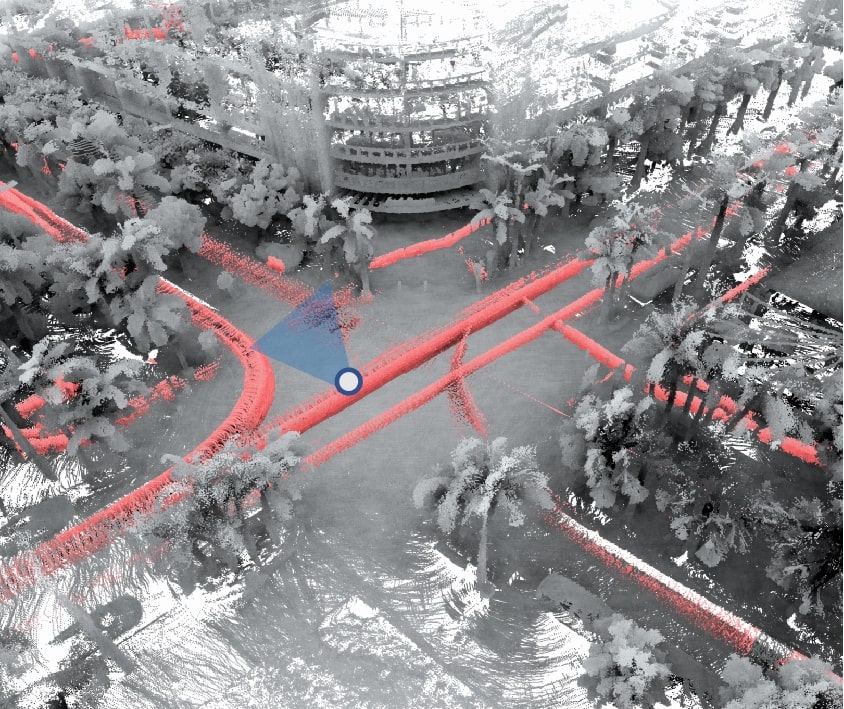}
    \caption{Ground truth}
    \label{subfig:av_gt}
    \vspace{0.5cm}
  \end{subfigure}
  \begin{subfigure}{0.195\textwidth}
    \centering
    \includegraphics[height=1.1in]{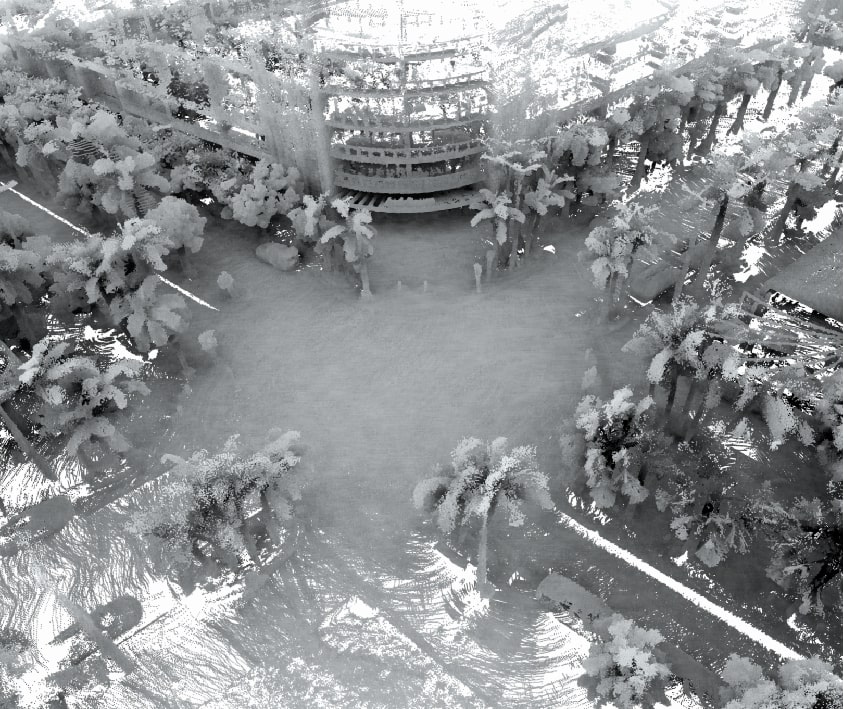}
    \caption{Ours}
    \label{subfig:av_ours}
    \vspace{0.5cm}
  \end{subfigure}
  \begin{subfigure}{0.195\textwidth}
    \centering
    \includegraphics[height=1.1in]{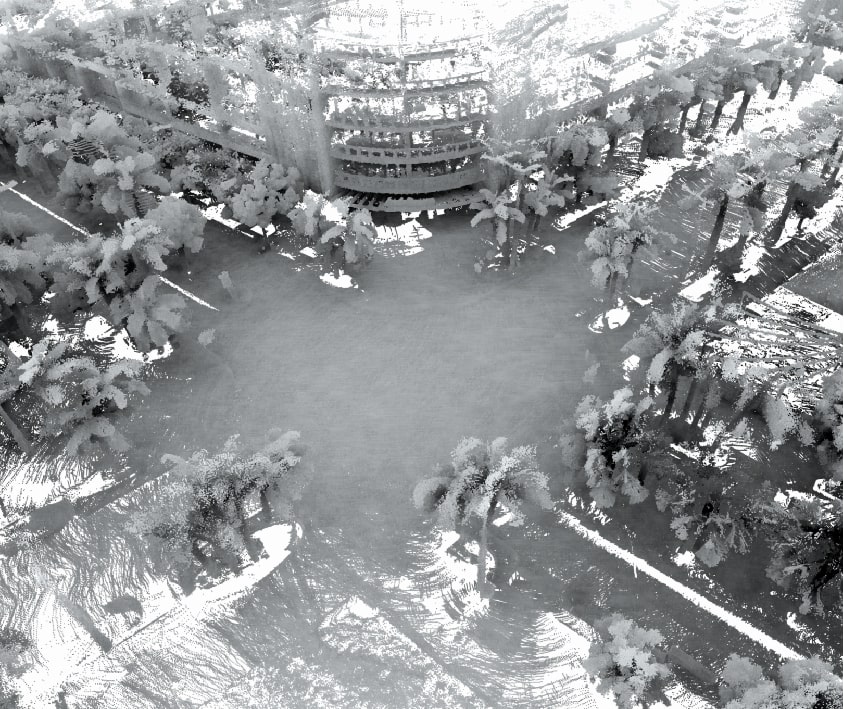}
    \caption{Erasor~\cite{lim2021ral}}
    \label{subfig:av_erasor}
    \vspace{0.5cm}
  \end{subfigure}
  \begin{subfigure}{0.195\textwidth}
    \centering
    \includegraphics[height=1.1in]{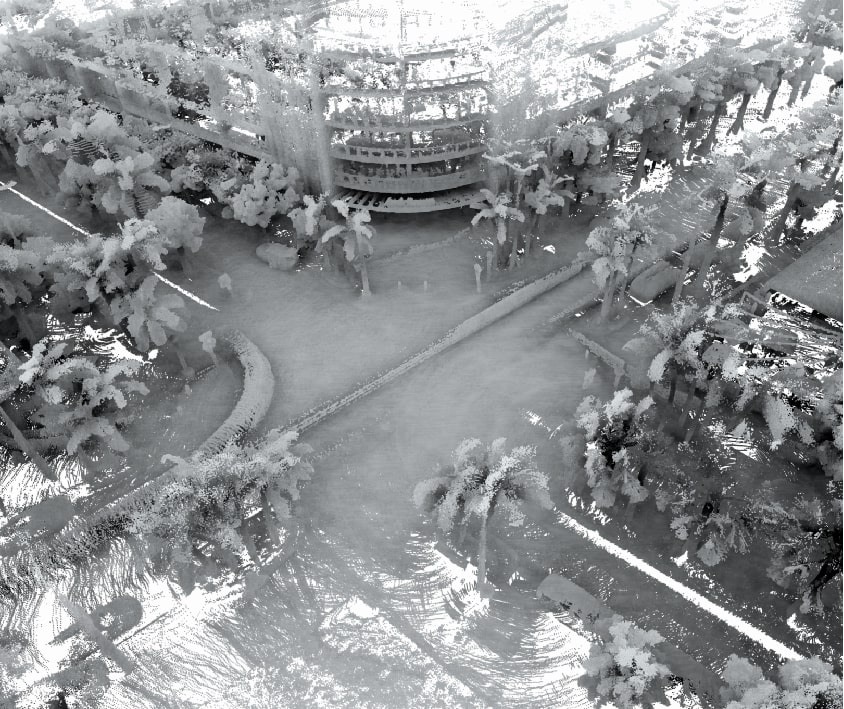}
    \caption{Removert~\cite{kim2020iros}}
    \label{subfig:av_removert}
    \vspace{0.5cm}
  \end{subfigure}
  \begin{subfigure}{0.195\textwidth}
    \centering
    \includegraphics[height=1.1in]{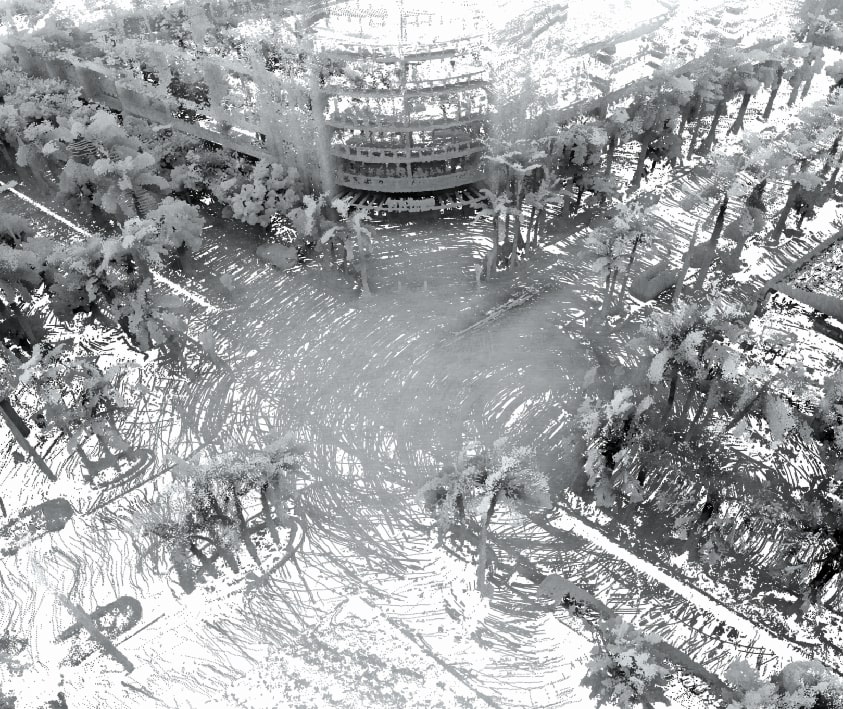}
    \caption{Octomap*~\cite{zhang2023ITSC}}
    \label{subfig:av_octomap}
    \vspace{0.5cm}
  \end{subfigure}
  \vspace{-4pt}
  \begin{subfigure}{0.195\textwidth}
    \centering
    \includegraphics[height=1.1in]{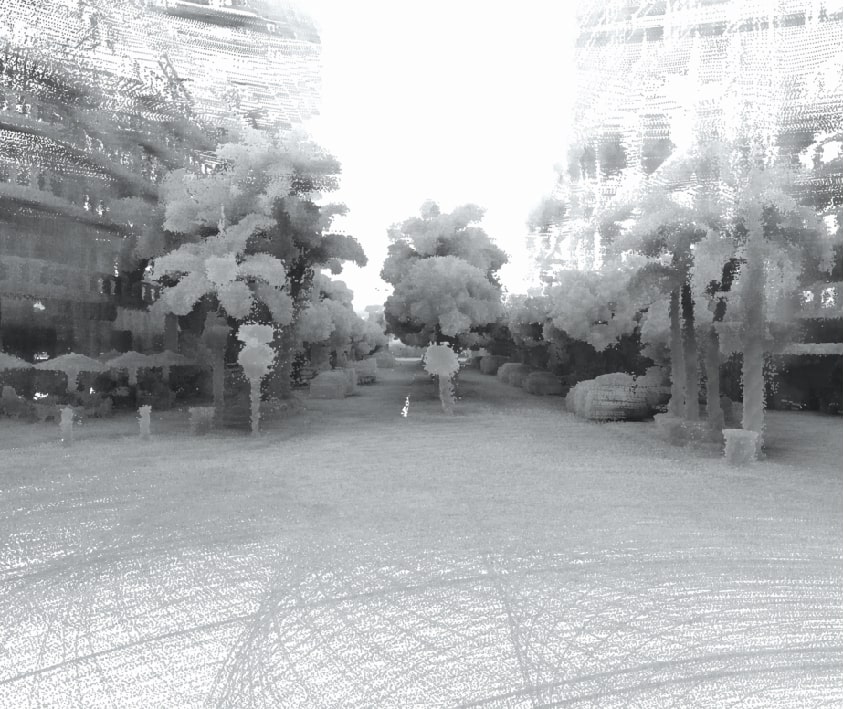}
    \caption{Ground truth}
    \label{subfig:av_gt_zoom}
  \end{subfigure}
  \begin{subfigure}{0.195\textwidth}
    \centering
    \includegraphics[height=1.1in]{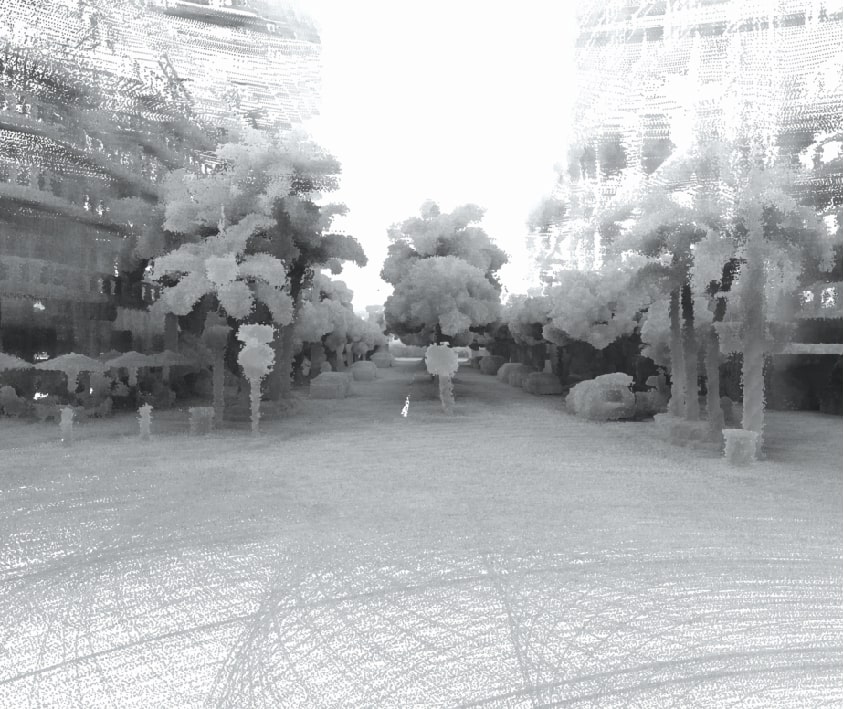}
    \caption{Ours}
    \label{subfig:av_ours_zoom}
  \end{subfigure}
  \begin{subfigure}{0.195\textwidth}
    \centering
    \includegraphics[height=1.1in]{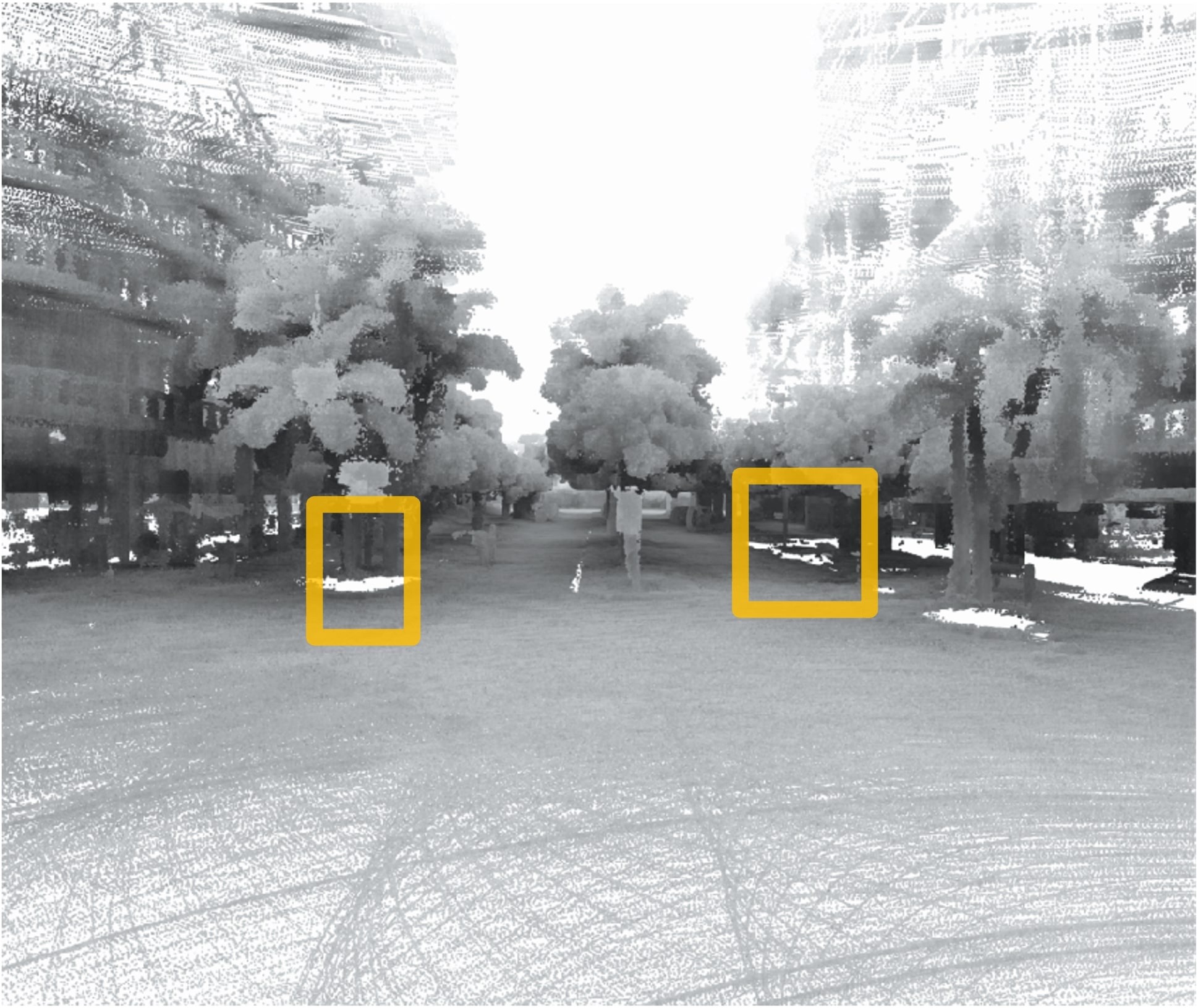}
    \caption{Erasor~\cite{lim2021ral}}
    \label{subfig:av_erasor_zoom}
  \end{subfigure}
  \begin{subfigure}{0.195\textwidth}
    \centering
    \includegraphics[height=1.1in]{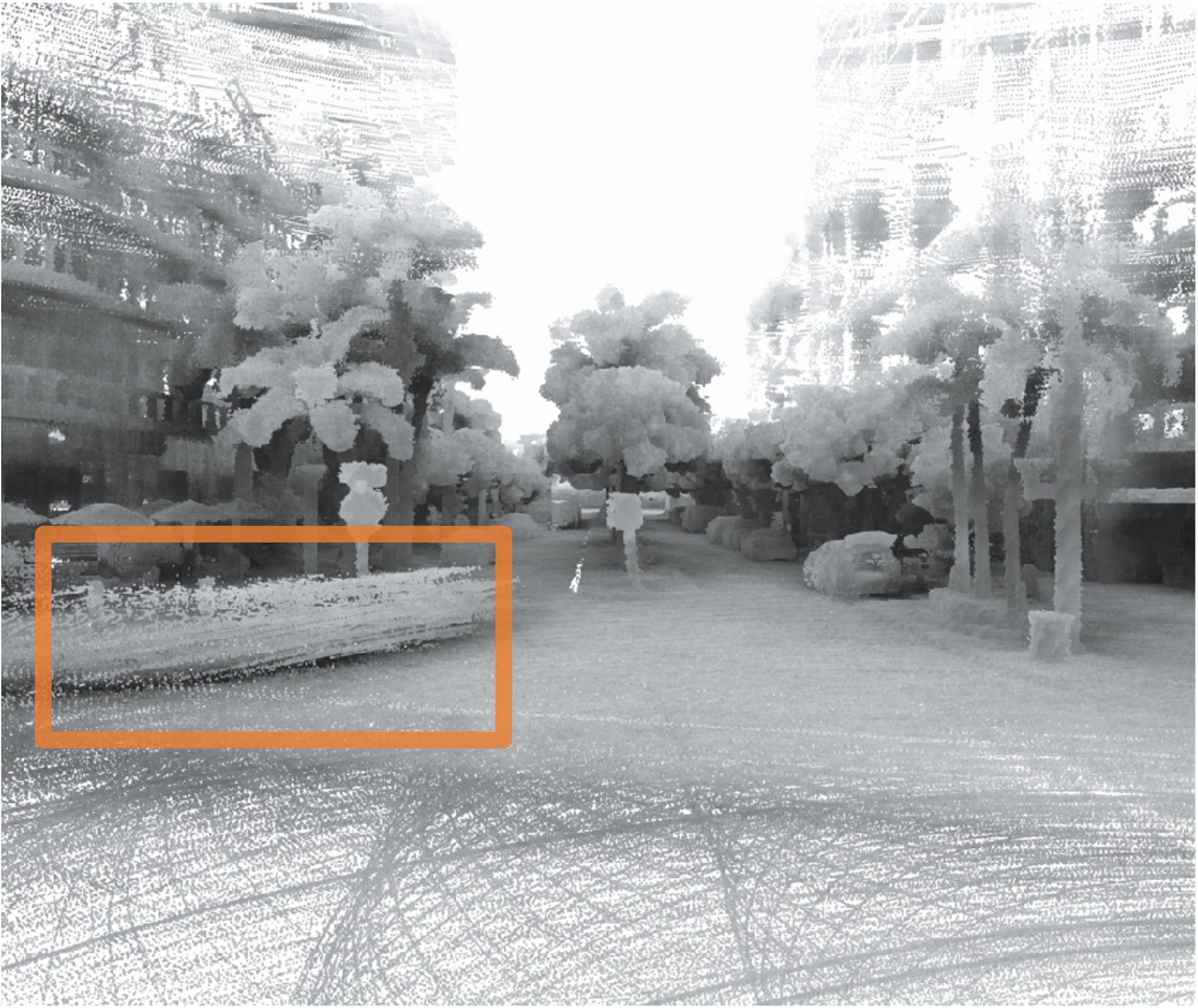}
    \caption{Removert~\cite{kim2020iros}}
    \label{subfig:av_removert_zoom}
  \end{subfigure}
  \begin{subfigure}{0.195\textwidth}
    \centering
    \includegraphics[height=1.1in]{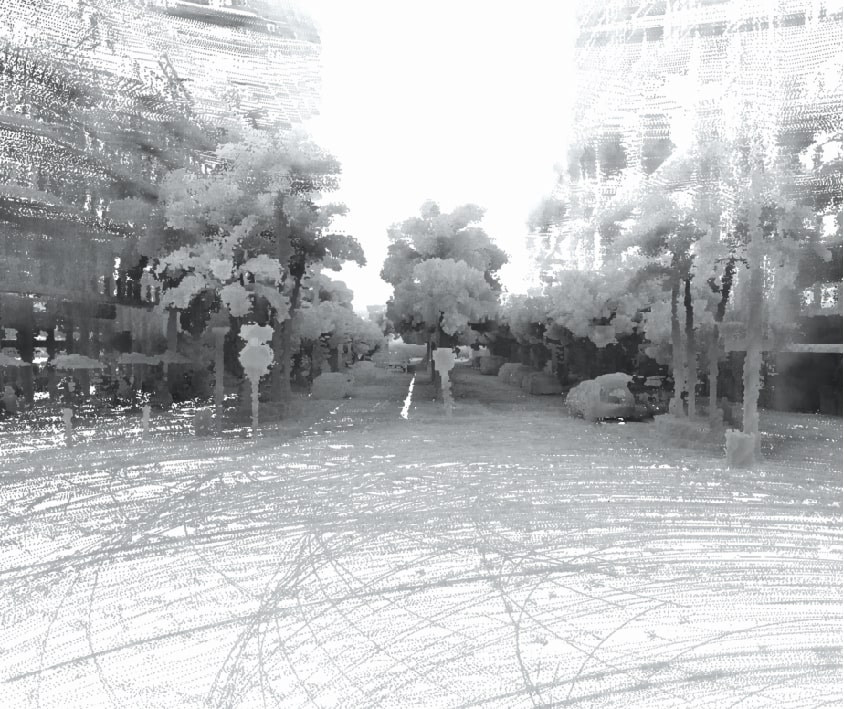}
    \caption{Octomap*~\cite{zhang2023ITSC}}
    \label{subfig:av_octomap_zoom}
  \end{subfigure}
  \vspace{-2pt}
  \caption{Comparison of dynamic object removal results produced by our proposed method and three baseline methods on the Argoverse2 data sequence of the KTH-benchmark.
    We show the bird's eye view on the first row and the zoomed view from the blue frustum shown in (a) on the second row. For the ground truth results in (a), the dynamic objects are shown in red.
    We only show the static points of ground truth for clearer comparison in zoomed view (f). We highlight the over-segmented parking car and sign by Erasor and the undetected moving vehicle by Removert.
    }
  \label{fig:av2_static_exp}
  \vspace{-8pt}
\end{figure*}

\subsection{Dynamic Object Segmentation}
\label{subsec:dynamicOS}

\textbf{Datasets.} For dynamic object segmentation, we use the KTH-Dynamic-Benchmark~\cite{zhang2023ITSC} for evaluation, which includes four sequences in total: sequence 00 (frame 4,390 -- 4,530 ) and sequence 05 (frame 2,350 -- 2,670) from the KITTI dataset~\cite{geiger2012cvpr,behley2019iccv},
which are captured by a 64-beam LiDAR, one sequence from the Argoverse2 dataset~\cite{wilson2021neurips} consisting of 575 frames captured
by two 32-beam LiDARs, and a semi-indoor sequence captured by a sparser 16-beam LiDAR. All sequences come with corresponding pose files and point-wise dynamic or static labels as the ground truth.
It is worth noting that the poses for KITTI 00 and 05 were obtained from SuMa~\cite{behley2018rss} and the pose files for the Semi-indoor sequence come from NDT-SLAM~\cite{saarinen2013iros-f3mi}.

\textbf{Metric and Baselines.}
The KTH-Dynamic-Benchmark evaluates the performance of the method by measuring the classification accuracy of dynamic points (DA\%),
static points (SA\%) and also their associated accuracy (AA\%) where $AA=\sqrt{DA\cdot SA } $.
The benchmark provides various baselines such as the state-of-the-art LiDAR dynamic object removal methods --
Erasor~\cite{lim2021ral} and Removert~\cite{kim2020iros},
as well as the traditional 3D mapping method, Octomap ~\cite{hornung2013ar,wurm2010icraws}, and its modified versions, Octomap
with ground fitting and outlier filtering.
As SHINE-mapping demonstrates the ability to remove dynamic objects in our static mapping experiments, we also report its result in this benchmark.
Additionally, we report the performance of the state-of-the-art online moving object segmentation methods, 4DMOS~\cite{mersch2022ral} and its extension MapMOS~\cite{mersch2023ral}.
As these two methods utilize KITTI sequences 00 and 05 for training, we only show the results of the remaining two sequences.
For the parameter setting, we set our method's leaf resolution to $0.3$\,m, and the threshold for segmentation as $d_\text{static} = 0.16\,$m.
We set the leaf resolution for Octomap to $0.1$\,m.

\textbf{Results.}
The quantitative results of the dynamic object segmentation are shown in \cref{tab:experiments_on_KTH}. And we depict the accumulated static points
generated by different methods in \cref{fig:av2_static_exp}.
We can see that our method achieves the best associated accuracy (AA) in three autonomous driving sequences (KITTI 00, KITTI 05, Argoverse2) and vastly outperforms baselines.
The supervised learning-based methods 4DMOS and MapMOS do not obtain good dynamic accuracy (DA) due to limited generalizability. Erasor and Octomap
tend to over-segment dynamic objects, resulting in poor static accuracy (SA). Removert and SHINE-mapping are too conservative and cannot detect all dynamic objects.
Benefiting from the continuity and large capacity of the 4D neural representation, we strike a better balance between preserving static background points and removing dynamic objects.

It is worth mentioning again that our method does not rely on any pre-processing or post-processing algorithm such as ground fitting,
outlier filtering, and clustering, but also does not require labels for training.

\section{Conclusion}
\label{subsec:conclusion}
In this paper, we propose a 4D implicit neural map representation for dynamic scenes that allows us to represent the TSDF of static and dynamic parts of a scene.
For this purpose, we use a hierarchical voxel-based feature representation that is then decoded into weights for basis functions to represent a time-varying TSDF that can be queried at arbitrary locations.
For learning the representation from a sequence of LiDAR scans, we design an effective data sampling strategy and loss functions.
Equipped with our proposed representation, we experimentally show that we are able to tackle the challenging problems of static mapping and dynamic object segmentation.
More specifically, our experiments show that our method has the ability to accurately
reconstruct 3D maps of the static parts of a scene and can completely remove moving objects at the same time.


\textbf{Limitations.}
While our method achieves compelling results, we have to acknowledge that we currently rely on estimated poses by a separate SLAM approach, but also cannot apply our approach in an online fashion. However, we see this as an avenue for future research into joint incremental mapping and pose estimation.

\textbf{Acknowledgements.}
We thank Benedikt Mersch for the fruitful discussion and for providing experiment baselines.


{
  \small
  \bibliographystyle{ieeenat_fullname}
  \bibliography{new,glorified}
}


\end{document}